\documentclass[letterpaper, 10 pt, conference]{ieeeconf}

\IEEEoverridecommandlockouts
\usepackage{graphics} 
\usepackage{amsmath}
\usepackage{amssymb}  
\usepackage{amsfonts} 
\usepackage{graphicx,float,wrapfig}
\usepackage{cite}
\usepackage{bm}
\usepackage{soul,color}
\usepackage[linesnumbered,ruled]{algorithm2e}
\usepackage{url}
\usepackage{booktabs}
\usepackage[table,xcdraw]{xcolor}
\usepackage{subcaption}

\usepackage{flushend}
\allowdisplaybreaks

\definecolor{ao(english)}{rgb}{0.0, 0.5, 0.0}

\newcommand{\bmat}{\begin{bmatrix}}
\newcommand{\emat}{\end{bmatrix}}
\newcommand{\argmin}{\text{argmin}}
\title{\LARGE \bf
Cooperation-Aware Lane Change Maneuver in Dense Traffic based on Model Predictive Control with Recurrent Neural Network
}

\author{Sangjae Bae$^{1*}$, Dhruv Saxena$^{2}$, Alireza Nakhaei$^{3}$, Chiho Choi$^{3}$, Kikuo Fujimura$^{3}$, and Scott Moura$^{1}$
\thanks{$^\ast$Corresponding author.}
\thanks{$^{1}$University of California, Berkeley, USA. {\tt\{sangjae.bae, smoura\}@berkeley.edu}}
\thanks{$^{2}$Carnegie Mellon University, USA. {\tt dsaxena@andrew.cmu.edu}}
\thanks{$^{3}$Honda Research Institute, USA. {\tt\{anakhaei, cchoi, kfujimura\}@honda-ri.com}}
\thanks{This work was funded by Honda Research Institute, USA.}}%

\begin{document}
\maketitle
\thispagestyle{empty}
\pagestyle{empty}

\begin{abstract}
This paper presents a \textit{real-time} lane change control framework of autonomous driving in dense traffic, which exploits cooperative behaviors of other drivers. This paper focuses on heavy traffic where vehicles cannot change lanes without cooperating with other drivers. In this case, classical robust controls may not apply since there is no ``safe'' area to merge to without interacting with the other drivers. That said, modeling complex and interactive human behaviors is highly non-trivial from the perspective of control engineers. We propose a mathematical control framework based on Model Predictive Control (MPC) encompassing a state-of-the-art Recurrent Neural network (RNN) architecture. In particular, RNN predicts interactive motions of other drivers in response to potential actions of the autonomous vehicle, which are then systematically evaluated in safety constraints. We also propose a \textit{real-time} heuristic algorithm to find locally optimal control inputs. Finally, quantitative and qualitative analysis on simulation studies are presented to illustrate the benefits of the proposed framework.
\end{abstract}

\section{Introduction}
An autonomous-driving vehicle is no longer a futuristic concept and extensive research have been conducted in various aspects, spanning from localization, perception, and controls, including implementation and validation. From the perspective of control engineering, designing a controller that ensures safety in various traffic conditions (e.g., arterial-roads, highways in free-flow/dense traffic, with/without traffic lights) has been a principal research focus. This paper focuses on lane change in dense traffic. 

Due to the importance of safety, many publications have focused on robust control that guarantee collision avoidance in the face of uncertainty. Unlike other autonomous robots, autonomous-driving vehicles can take advantage of existing roadway infrastructure, such as arterial roads and highways. By exploiting the roadway for guidance, longitudinal control designs have proven their effectiveness in maintaining a safety distance to a front vehicle, as well as maintaining driving comfort and energy efficiency \cite{bae2019design,bae2019real}. A lane-changing controller extends longitudinal control without significant mathematical modifications by recognizing additional lanes and vehicles \cite{hatipoglu2003automated}. Lane-changing controllers often utilize probabilistic models \cite{vitus2013probabilistic, ulbrich2013probabilistic} or scenario-based models \cite{schildbach2015scenario,chandra2017safe} to predict adjacent drivers' motions and intentions, which primarily define a safe area in a robust manner. 

\begin{figure}
    \centering
    \includegraphics[width=0.8\columnwidth]{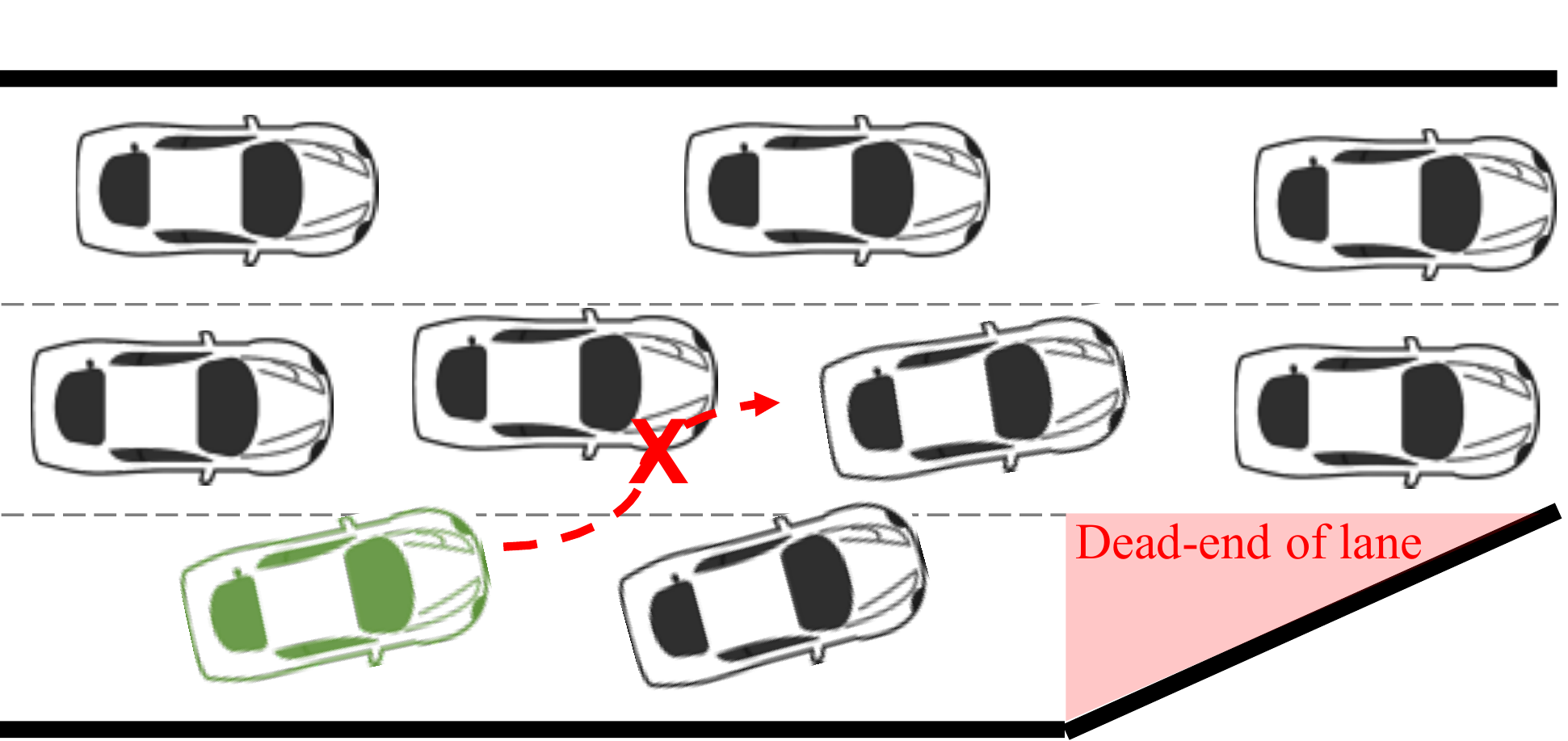}
    \caption{\small{The autonomous-driving vehicle (in \textcolor{ao(english)}{green}) intends to change lanes, within a restricted merging area. The traffic is dense with minimal distance intervals between vehicles and any interval is spatially insufficient for the autonomous-driving vehicle to merge into. The autonomous-driving vehicle would get stuck in the merging area, unless other vehicles slow down to make a space for the vehicle. Cooperative behaviors of other vehicles are obscure to the autonomous-driving vehicle.}}
    \label{fig:motivation}
\end{figure}

Unfortunately, those robust methods may not apply in highly dense traffic where there is no ``safe'' area to merge into (see Fig.~\ref{fig:motivation}). In dense traffic conditions, interactions with other drivers are essential to successfully change lanes by leveraging cooperative behavior. That is, drivers can potentially slow down to create a spatial interval so that another vehicle can merge in. That said, modeling interactive behaviors by formal statistical or scenario-based approaches is highly non-trivial, due to its complex and stochastic nature as well as computational challenges. Consequently, lane changing that exploits cooperation with other drivers remains as an open research question. Nonetheless, solving this problem is critical to realizing fully autonomous-driving vehicles \cite{levinson2011towards}.

There is a small yet rich body of literature focused on human interactions during lane change or merging \cite{bouton2019cooperation, naranjo2008lane, you2015trajectory, sadigh2016planning, hu2019IDAS}. These methods limit their focus on interactions between \textit{two} vehicles at a time, in lane changing or lane keeping. In more realistic settings of highly dense traffic, each vehicle's motion is reactive to multiple vehicles simultaneously, and thus existing methods may not fit. Recently, Reinforcement Learning (RL) techniques have been thoroughly investigated. RL methods are appealing for their potential in finding maneuvers under interactive or unknown traffic conditions \cite{sadigh2016planning, shalev2016safe, saxena2019driving}. However, safety, reliability, and/or interpretability limit RL's practical use when human safety is at stake, and theses issues remain as active research areas \cite{garcia2015comprehensive, isele2018safe}. 



That being said, an increasing body of literature has applied Deep Neural Network architectures for autonomous driving and Advanced Driving Assistance Systems (ADAS). Deep neural networks have proven useful in explaining complex environments \cite{tian2018deeptest, chen2015deepdriving, bojarski2017explaining}. Recurrent Neural Network (RNN) architectures have been particularly effective in predicting motions of human (drivers), with respect to both accuracy \cite{choi2019drogon,alahi2016social} and computational efficiency \cite{gupta2018social}. Therefore, it would be natural to take advantage of those advances for controller design, yet still leverage rigorous control theory \cite{doyle2013feedback} and established vehicle dynamics models \cite{Kong2015} to yield safety guarantees. That is, we seek a controller that embeds accurate predictions of human interaction via RNNs, yet still maintains safety guarantees afforded by control theory. The incorporation of these two elements would yield a controller that is reliable, interpretable, and tunable, while containing a data-driven model that captures interactive motions between drivers. It is still challenging to mathematically incorporate RNN into formal controller design, and to solve the corresponding control problem effectively and efficiently. We address these challenges in this paper.

We add two original contributions to the literature: (i) We propose a mathematical control framework that systematically evaluates several other drivers' interactive motions, in highly dense traffic on the highway. The framework partially exploits Recurrent Neural networks (RNN) to predict (nonlinear) interactive motions, which are incorporated as safety constraints. Optimal control inputs are obtained via Model Predictive Control (MPC) based on vehicle dynamics. (ii) We propose a \textit{real-time} rollout-based heuristic algorithm that sequentially evaluates other driver's reactions and finds locally optimal solutions. The idea of incorporating RNN as a prediction model into a MPC controller is straightforward. However, a problem with RNN becomes is the complexity and nonlinearities provides challenges for classical optimization algorithms. We show our heuristic algorithm finds locally optimal solutions effectively and efficiently.

The paper is organized in the following manner. Section \ref{sec:design_online_controller} describes the mathematical formulation of the proposed control framework, and the heuristic algorithm. Section \ref{sec:simulation} presents and analyzes simulation results to validate the proposed framework. Section \ref{sec:conclusion} summarizes the paper's contributions. 

\section{Controller Design}\label{sec:design_online_controller}
\begin{figure}
    \centering
    \includegraphics[width=1\columnwidth]{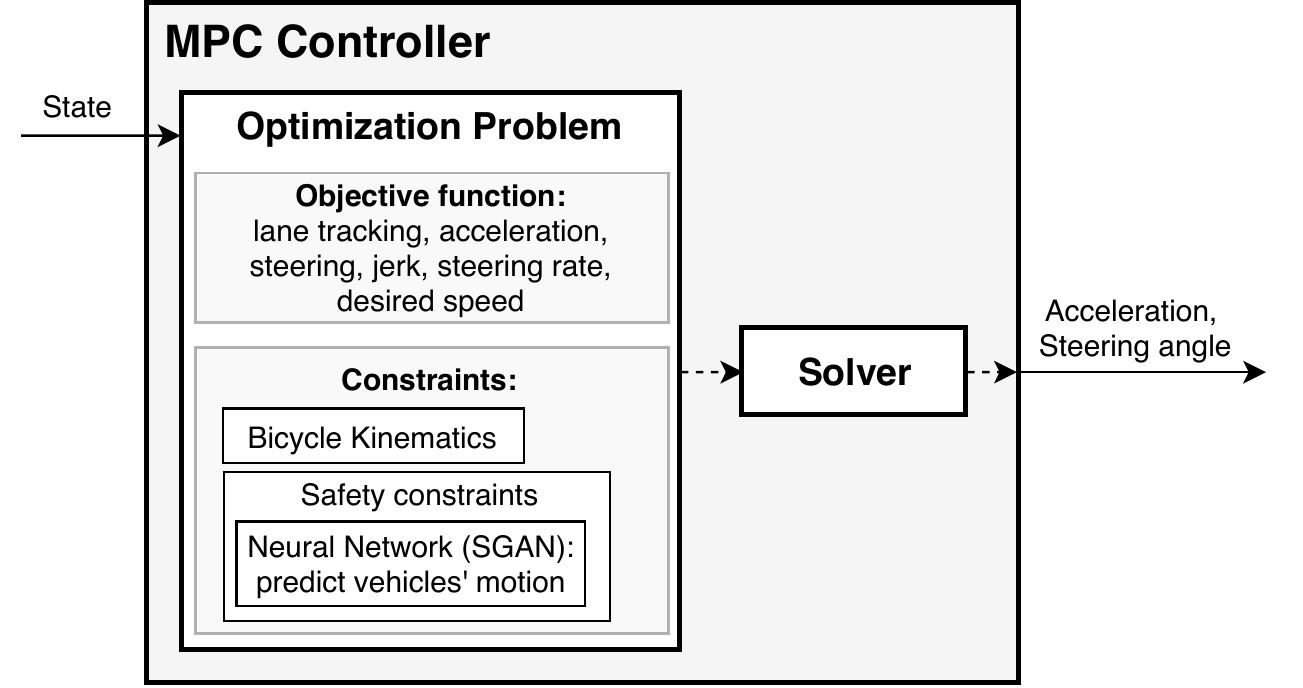}
    \caption{\small{Diagram of the control framework with a Recurrent Neural network, SGAN. The solver (small box in MPC controller) can be based on different algorithms.}}
    \label{fig:diag}
\end{figure}

To design a lane changing controller for highly dense traffic, it is critical to estimate how surrounding vehicles will react to our vehicle. Predicting their behaviors, however, is challenging due to their interactive nature, and it is complex to model either physically or statistically, using formal methods such as Markov chains. We therefore employ a Recurrent Neural Network architecture (RNN), particularly a Social Generative Adversarial Network (SGAN) that has been successful in predicting interactive human behaviors in crowded spaces \cite{gupta2018social}.
Fig.~\ref{fig:diag} shows a schematic of the control framework. The key intuition of the framework is the following. The controller uses Model Predictive Control (MPC) as the basis, which takes current states (position, velocity, and heading) as inputs and provides a pair of acceleration and steering angle as output. In the MPC controller, the optmization objective is to track a target lane, a desired speed, and penalize large control efforts, while satisfying system dynamics and avoiding collisions. The solver, based on a heuristic algorithm, solves the optimization problem at each time step by exploiting a trained SGAN to predict the other vehicles' interactive motions that are reactive to the ego vehicle's \footnote{We refer to ``ego vehicle'' as the vehicle that is controlled.} actions. Note that the predicted motions are incorporated into the constraint of collision avoidance. The controller design is detailed in the following sections.

\subsection{System Dynamics}\label{ssec:system_dynamics}
We utilize the nonlinear kinematic bicycle model in \cite{Kong2015} to represent the vehicle dynamics. For completeness, we re-write the kinematics here:
\begin{align}
    \dot{x}&=v\cos(\psi+\beta)\\
    \dot{y}&=v\sin(\psi+\beta)\\
    \dot{\psi}&=\frac{v}{l_r}\sin(\beta)\\
    \dot{v}&=a\\
    \beta&=\tan^{-1}\left(\frac{l_r}{l_f+l_r}\tan(\delta)\right)
\end{align}
where ($x$,$y$) is the Cartesian coordinate for the center of vehicle, $\psi$ is the inertial heading, $v$ is the vehicle speed, $a$ is the acceleration of the car's center in the same direction as the velocity, and $l_f$ and $l_r$ indicate the distance from the center of the car to the front axles and and to the rear axles, respectively. The control inputs are: (front wheel) steering angle $\delta$ and acceleration $a$. We use Euler discretization to obtain a discrete-time dynamical model in the form:
\begin{equation}
    z(t+1) = f(z(t),u(t)),
\end{equation}
where $z = \begin{bmatrix}x&y&\psi&v\end{bmatrix}^\top$ and $u = \begin{bmatrix}a&\delta\end{bmatrix}^\top$.

\subsection{Control Objective}
The control objective is to merge to target lane, while avoiding collisions with other vehicles. We prefer to change lanes sooner. We also prefer smooth accelerations and steering for drive comfort. 
The objective function is formulated:
\begin{align}
    J&=\sum_{\ell=t}^{t+T}\lambda_{div}(x(\ell|t);x_{\text{end}})D(\ell|t)
    \label{eq:obj_divergence}\\
     &+\sum_{\ell=t}^{t+T}\lambda_{v}\|v(\ell|t)-v^{\text{ref}}\|^2\\
     &+\sum_{\ell=t}^{t+T-1}\lambda_{\delta}\|\delta(\ell|t)\|^2 \label{eq:obj_front_wheel_angle}\\
     &+\sum_{\ell=t}^{t+T-1}\lambda_{a}\|a(\ell|t)\|^2\label{eq:obj_acc}\\
     &+\sum_{\ell=t+1}^{t+T-1}\lambda_{\Delta\delta}\|\delta(\ell|t)-\delta(\ell-1|t)\|^2\label{eq:obj_front_wheel_angle_jerk}\\
     &+\sum_{\ell=t+1}^{t+T-1}\lambda_{\Delta a}\|a(\ell|t)-a(\ell-1|t)\|^2\label{eq:obj_jerk}
\end{align}
where $z(\ell|t)$ and $u(\ell|t)$ are, respectively, states and input at time $\ell$ based on the measurements at time $t$. Symbol $x_\text{end}$ is the latitude coordinate of the road-end, $D(\ell|t)$ is the distance norm for the vector between the ego vehicle's center and the target lane at time $\ell$, $v^{\text{ref}}$ is the reference velocity. Each penalty is regularized with $\lambda_{div}$, $\lambda_v$, $\lambda_{\delta}$, $\lambda_{a}$, $\lambda_{\Delta \delta}$, and $\lambda_{\Delta a}$, respectively. We incentivize a timely lane change with the dynamic weight $\lambda_{div}$, written as a convex-function, $\lambda_{div} = \|\frac{1}{x_\text{end}-x}\|$.
The term \eqref{eq:obj_divergence} penalizes the divergence of the center of the vehicle from the vertical center of the target lane. The term \eqref{eq:obj_front_wheel_angle} and \eqref{eq:obj_acc} penalize the control effort of steering angle and acceleration, respectively. The term \eqref{eq:obj_front_wheel_angle_jerk} and \eqref{eq:obj_jerk} penalize the steering rate and jerk, respectively, for driver comfort. 


\subsection{Constraints for Safety with a Recurrent Neural Network}
To quantify safety, we consider a distance measure between two vehicles. That is, if distance between two vehicles is zero, it means they collide into each other. A mathematical measure of distance between two vehicles depends on how each vehicle is shaped, which we discuss next.

\subsubsection{Vehicle Shape Model}
\begin{figure}
    \centering
    \includegraphics[width=0.4\columnwidth]{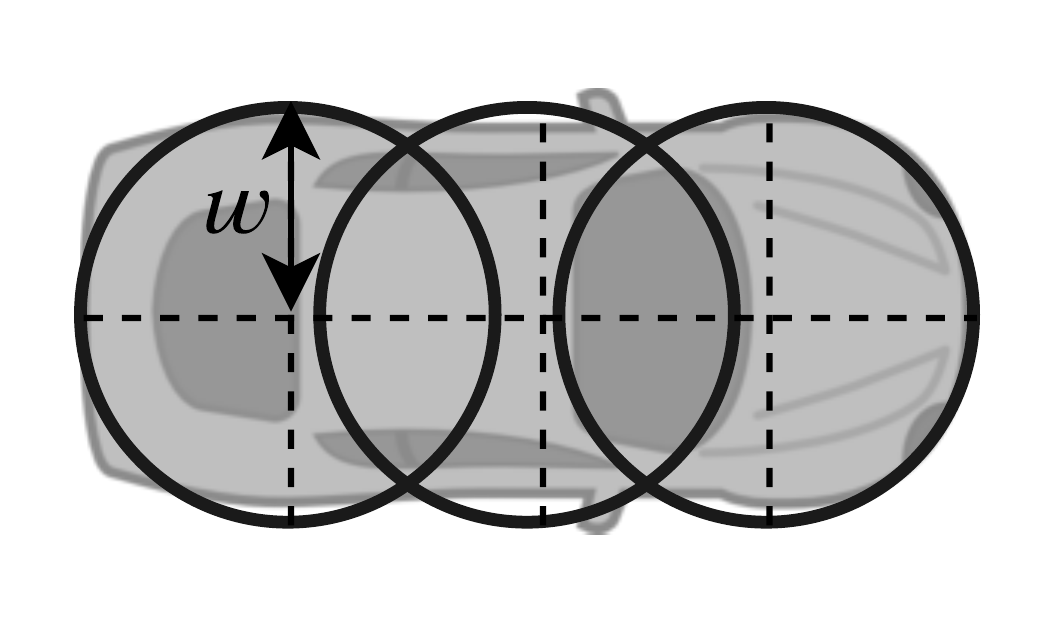}
    \caption{\small{Vehicle shape modeled by three circles.}}
    \label{fig:vehicle_shape}
\end{figure}

We aim to model the vehicle shape such that it admits a smooth distance measure that is continuously differentiable, to enable gradient-based algorithms during optimization. We model it with three circles as illustrated in Fig.~\ref{fig:vehicle_shape}. 
With the circle model, the safety constraint with respect to vehicle $i$ is written as:
\begin{equation}
     g_i(x,y)=\min_{p,q\in\{-1,0,1\}} d_{i}(p,q) \geq \epsilon,\label{eq:const_circle}
\end{equation}
\begin{align}
    &\text{where: } d_{i}(p,q)\nonumber\\ 
    &= \big((x+p(h-w)\cos\psi)-(x_i+q(h_i-w_i)\cos\psi_i)\big)^2\nonumber\\
    &\;\;+\big((y+p(h-w)\sin\psi)-(y_i+q(h_i-w_i)\sin\psi_i)\big)^2\nonumber\\
    &\;\;-(w+w_i)^2\label{eq:dist_circle}
\end{align}
and $w$ and $h$ are, respectively, half width and half height of the vehicle, respectively, and $\epsilon$ is a safety bound. 

\subsubsection{Interactive Motion Prediction}
Drivers' motions are responsive to interactions with each other, and consequently the motions must be predicted simultaneously. We utilize social GAN (SGAN) in \cite{gupta2018social} which efficiently and effectively captures multi-modal interactions between agents (drivers). SGAN is composed of a generator and discriminator that are adversarial to each other. 
Both the generator and discriminator are comprised of long-short term memory (LSTM) networks to account for the sequential nature of agents' motion. 
The following module pools the motion states of agents with each other to evaluate their interactions. This pooling process is essentially to share the motion history between agents, generating social interactions as a pooled tensor $P_i$ for \textit{each} agent $i$. The decoder in the generator then predicts multiple trajectories that are socially interactive with each other. 
SGAN takes as an input the sequence of positions for each agent over an observation time horizon $T_{obs}$. It outputs a sequence of positions for each agent over the prediction time horizon $T_{pred}$. Interested readers are referred to \cite{gupta2018social} for more details. It is important to highlight that a trained SGAN will predict the most probable reactions of other vehicles based on the control commands on the ego vehicle, and the history of previous actions. However, in reality, the reactions of other vehicles might be different.





\subsubsection{Incorporation of Recurrent Neural Network into Safety Constraints}
A trained SGAN predicts coordinates for the center of vehicle $i$. These coordinates are incorporated into safety constraints \eqref{eq:const_circle}. To formalize, we denote a trained SGAN as a function $\phi(t)$ that maps observed trajectories to predicted trajectories:
\begin{align}
    &\phi(t)\;:\;\nonumber\\ 
    &\begin{bmatrix}
    (x_1(t),y_1(t))&\cdots&(x_{N}(t),y_{N}(t))\\
    \vdots&\vdots&\vdots\\
    \begin{matrix}(x_1(t-T_{obs}+1),\\\qquad y_1(t-T_{obs}+1))\end{matrix} &\cdots&\begin{matrix}(x_{N}(t-T_{obs}+1),\\\qquad y_{N}(t-T_{obs}+1))\end{matrix}
    \end{bmatrix}\nonumber\\
    &\vspace{3mm}\hspace{4cm}\downarrow\vspace{3mm}\nonumber\\
    &\quad\begin{bmatrix}
    (\hat{x}_1(t+1),\hat{y}_1(t+1))&\cdots&(\hat{x}_{N}(t+1),\hat{y}_{N}(t+1))\\
    \vdots&\cdots&\vdots\\
    \begin{matrix}(\hat{x}_1(t+T_{pred}),\\\qquad \hat{y}_1(t+T_{pred}))\end{matrix}&\cdots&\begin{matrix}(\hat{x}_{N}(t+T_{pred}),\\\qquad\hat{y}_{N}(t+T_{pred}))\end{matrix}
    \end{bmatrix},
\end{align}
where  
$\hat{\cdot}$ indicates a predicted value. Given the observations until time $t$, the coordinates of vehicle $i$ at time $t+1$ are represented as $\hat{x}_{i}(t+1) = \phi_{i,x}(t)$ and $\hat{y}_{i}(t+1) = \phi_{i,y}(t)$.

\subsection{Control Problem Formulation}
The complete optimization problem for the receding horizon control is:
\begin{align}
    \min_{z,a,\delta}\;\; J&=\sum_{\ell=t}^{t+T}\Big(\lambda_{div}(x(\ell|t);x_{\text{end}})D(\ell|t)\nonumber\\
     &\quad\quad\quad+\lambda_v\|v(\ell|t)-v^{\text{ref}}\|^2\Big)\nonumber\\
     &+\sum_{\ell=t}^{t+T-1}\Big(\lambda_{\delta}\|\delta(\ell|t)\|^2+\lambda_{a}\|a(\ell|t)\|^2\Big)\nonumber\\
     &+\sum_{\ell=t+1}^{t+T-1}\Big(\lambda_{\Delta\delta}\|\delta(\ell|t)-\delta(\ell-1|t)\|^2\nonumber\\
     &\quad\quad\quad+\lambda_{\Delta a}\|a(\ell|t)-a(\ell-1|t)\|^2\Big)\label{eq:obj}
\end{align}
subject to:
\begin{align}
    z(\ell+1|t)&=f(z(\ell|t),\delta(\ell|t),a(\ell|t))\label{eq:const_dynamics}\\
    g_i (z(\ell+1|t)) &\geq \epsilon, \quad\forall i \in \{1,\cdots,N_{veh}\}\label{eq:const_collision}\\
    \delta(\ell|t) &\in [\delta_{\min},\delta_{\max}]\label{eq:const_delta}\\
    a(\ell|t) &\in [a_{\min},a_{\max}]\label{eq:const_a}\\
    x(\ell+1|t) &\leq x_\text{end} \label{eq:const_x_end}
\end{align}
where $\delta_{\min}$ and $\delta_{\max}$ are the minimum and maximum steering angle rate, respectively, and $a_{\min}$ and $a_{\max}$ are the minimum and maximum acceleration, respectively. The constraint \eqref{eq:const_x_end} ensures that the ego vehicle changes lanes before the road-end $x_\text{end}$. Note that the constraints \eqref{eq:const_dynamics} and \eqref{eq:const_collision} are nonlinear, and therefore the problem is non-convex. 

\subsection{Heuristic Algorithm}
Note that the inequality constraint \eqref{eq:const_collision} does not form a convex set. We assume that the mathematical form of the SGAN is not known, i.e. it is black-box, and we can only evaluate its output given an input. Along with the nonlinear equality constraint \eqref{eq:const_dynamics}, it is non-trivial to find a solution with canonical optimization algorithms, such as gradient descent or Newton methods. We propose a rollout-based algorithm that finds locally optimal solutions in a time efficient manner. The process is as follows. From the current state at time $t$, the controller randomly generates a finite set of control sequences over the time horizon $T$. That is, $\mathcal{U} = \{U_1, \cdots,U_j,\cdots, U_{N_{\text{sim}}}\}$ where $U_j=\bmat u_j(0|t)&\cdots&u_j(T-1|t)\emat^\top$. Then the controller evaluates the cumulative cost \eqref{eq:obj} for each control sequence and chooses one that has a minimum cost over the time horizon $T$. The controller takes the first control input from the optimal sequence and discards the rest. Algorithm \ref{alg:monte_rollout} formalizes the procedure. 

The process of finding the optimal sequence (line \ref{alg:eval_cost} in Algorithm \ref{alg:monte_rollout}) is the following. At each time step $\ell$, consider the $j^{th}$ candidate control sequence $U_j$. The controller (i) runs SGAN to predict the motions of surrounding vehicles, (ii) it propagates the ego vehicle through the dynamics \eqref{eq:const_dynamics} given the action $U_j$, 
(iii) it checks the constraints \eqref{eq:const_collision}-\eqref{eq:const_x_end}, and discards the candidate sequence if the constraints are violated, and (iv) it updates the cumulative cost \eqref{eq:obj}. Note that evaluating each control sequence candidate can be parallelized, and therefore a parallel computation framework can be applied to improve computational efficiency.


\begin{algorithm}[t]
    \SetKwInOut{Input}{Input}
    \SetKwInOut{Output}{Output}
    \SetKwInOut{Init}{Init}
    \Init{states $z = z_0$, \\
    other vehicles' position $(x_i,y_i) = (x_{i0},y_{i0})$ for all $i\in\{1,\cdots,N_{veh}\}$}
    \While{$x < x_{\text{end}}$ and $D \neq 0$}{
        Randomly generate a total of $N_{\text{sim}}$ control sequences over $T$ while satisfying \eqref{eq:const_delta}, \eqref{eq:const_a}\\
        $\mathcal{U}=\{U_1, \cdots,U_j,\cdots, U_{N_{\text{sim}}}\}$
        \vspace{3mm}
        
        Find the optimal sequence that minimizes cumulative cost over $T$ and that is feasible with \eqref{eq:const_collision}\\
        $U_\ast \leftarrow \argmin_{U\in\mathcal{U}}\; \eqref{eq:obj}$ \label{alg:eval_cost}\\
        \vspace{3mm}
        
        Propagate through dynamics \eqref{eq:const_dynamics} with the first element of $U_\ast$\\
        $z \leftarrow f(z,[U_\ast]_0)$\label{alg:monte_update_state}\\
        \vspace{3mm}
        
        Observe positions of other vehicles at the current time $t$\\
        $(x_i,y_i) \leftarrow (x_i(t),y_i(t))$ for all $i$
    }
    \caption{Monte Carlo Roll-out Algorithm}\label{alg:monte_rollout}
\end{algorithm}

Algorithm \ref{alg:monte_rollout} is straightforward to implement, however, it may require substantial computation power to find a solution in real-time, depending on the time horizon $T$ and the sample size $N_{\text{sim}}$. That said, unlike other applications of motion planning algorithms, autonomous driving on roads has specific patterns in terms of actions in specific driving scenarios. For example, if a vehicle keeps driving in the same lane, large steering angles do not have to be explored. Similarly, if a vehicle changes lane to the left, then steering angles to right do not have to be explored. From those observations, a smaller size of action spaces can be specified in each driving scenario: (i) keeping lane, (ii) changing lane to the left, and (iii) changing lane to the right. Denote the action spaces of lane keeping by $\mathcal{A}_{M}$, changing lane to the left by $\mathcal{A}_{L}$, and changing lane to the right by $\mathcal{A}_{R}$. Then each action space can read:
\begin{align}
    \mathcal{A}_{M} &= [a_\text{min}, a_\text{max}]\times[\alpha\delta_\text{min}, \alpha\delta_\text{max}],\\
    \mathcal{A}_{L} &= [a_\text{min}, a_\text{max}]\times[0, \delta_\text{max}],\label{eq:action_space_left}\\
    \mathcal{A}_{R} &= [a_\text{min}, a_\text{max}]\times[\delta_\text{min}, 0],\label{eq:action_space_right}
\end{align}
where $\alpha \in [0,1]$, $\delta>0$ indicates steering to the left, and $\delta<0$ indicates steering to the right. The action space 
\eqref{eq:action_space_left} indicates that we only consider steering angles to the left when turning left, and the same logic is applied to the action space 
\eqref{eq:action_space_right}.

\section{Simulation Study}
\label{sec:simulation}
\subsection{Driver Model of Cooperativeness} \label{ssec:driver_model}
\begin{figure}
    \centering
    \includegraphics[width=.8\columnwidth,trim={0 1.2cm 0 1.1cm},clip]{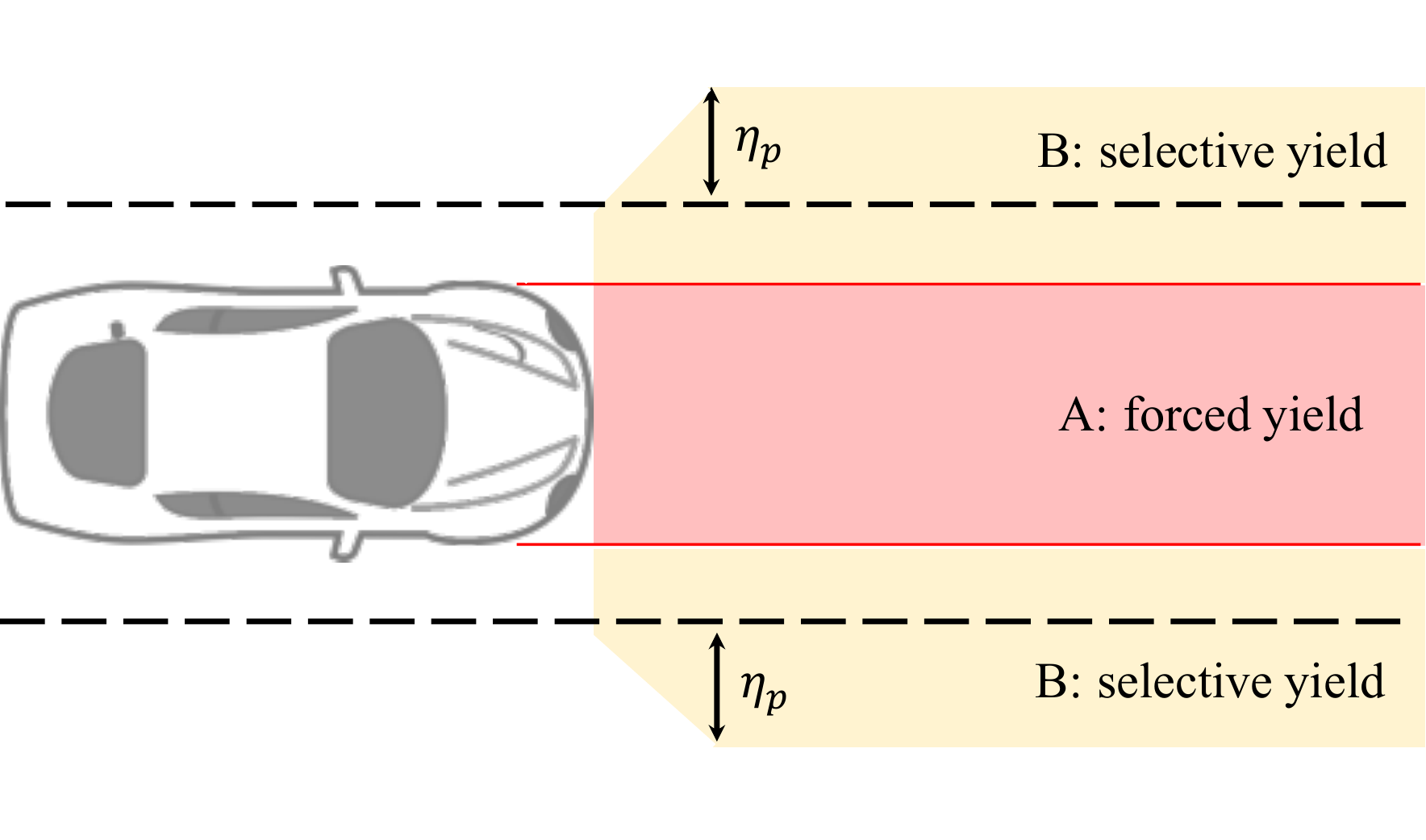}
    \caption{\small{Forced yield zone (red area labeled with A) and selective yield zone (yellow area labeled with B). The dashed lines indicate the boundaries of the center lane. If another vehicle from the next lane intersects with the path of the vehicle, i.e. in zone A, the vehicle in the center lane must stop and wait until the other vehicle cuts into the center lane. If the other vehicle from the next lane intersects with zone B, then the vehicle in the center lane decides to either yield or not, according to the cooperativeness parameter $\eta_{c}$. Zone B corresponds to a vertical perception range, and the range can be adjusted by $\eta_{\textit{p}}$.}}
    \label{fig:driver_model_yield_zone}
\end{figure}

For the simulation study, we have a driver model in which the longitudinal dynamics is controlled by an intelligent driver model (IDM) from \cite{treiber2000congested}. The lane changing behavior is governed by the strategy of Minimizing Overall Braking Induced by Lane changes (MOBIL) from \cite{kesting2007general}. The driver model is also based on the bicycle kinematics in Section \ref{ssec:system_dynamics}. Additionally, we introduce a parameter for cooperativeness $\eta_{c}\in [0,1]$ to the driver model. If $\eta_{c} = 1$, then a vehicle stops and waits until another vehicle within the selective yield zone (zone B in Fig.~\ref{fig:driver_model_yield_zone}) overtakes. If $\eta_{c} = 0$, then the vehicle ignores neighboring vehicles and drives forward. If $0 < \eta_{c} < 1$, then the yield action is randomly sampled from the Bernoulli distribution with probability $p=\eta_{c}$.

\subsection{Simulation Scenario Overview}
\begin{figure}
    \centering
    \includegraphics[width=1\columnwidth]{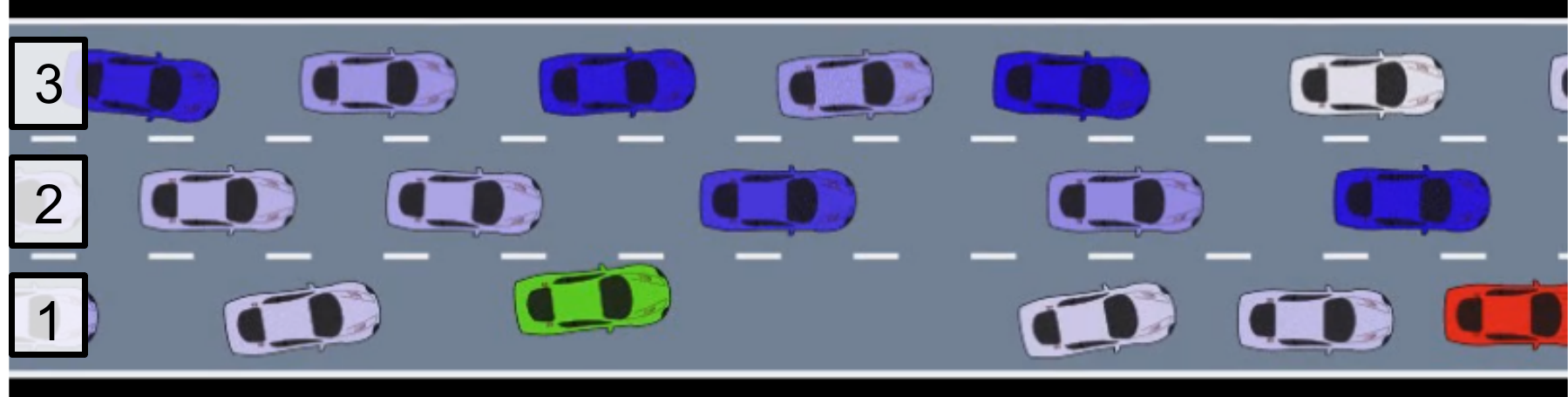}
    \caption{\small{Merging scenario test setup with three lanes. Green indicates the ego vehicle, red indicates a stopped vehicle, i.e. dead-end of the lane, and the shades of blue indicate other drivers. The lighter blue corresponds to less cooperative drivers, i.e. $\eta_{c}$ near 0. Each number in the white rectangle indicates the lane number.}}
    \label{fig:merging_scene}
\end{figure}

In the simulation study, we consider a highway segment with dense traffic, illustrated in Fig.~\ref{fig:merging_scene}. In the scene, the ego vehicle (in green) plans to change lanes from the current lane (lane 1) to the next lane (lane 2). However, the vehicles are driving with narrow distance intervals between each other, which are not large enough for the ego vehicle to cut in without cooperating with other vehicles (in various shades of blue). Given this challenge, the controller seeks a pair of acceleration and steering angle trajectories to induce the other vehicles to make space for the ego vehicle to cut in.

Only the ego vehicle uses the controller designed in Section \ref{sec:design_online_controller} and the other vehicles (in blue) are based on the driver model in the previous Section \ref{ssec:driver_model} with heterogeneous parameter settings. The controller design parameters are such that the divergence from the target lane is more significantly penalized relative to the other penalty terms. The list of controller design parameters is detailed in Table \ref{table:control_params}. Note that it is preferred to have a receding time horizon $T$ that is sufficiently long to account for other drivers' reaction time. 

The driver model parameters are set in the following manner: (i) The minimum distance between vehicles is shorter than the vehicle length, which creates the highly dense traffic we are interested in. (ii) The other drivers drive realistically, i.e. they apply physically feasible acceleration and steering angles. (iii) The other drivers' motions are noisy, i.e. their center positions oscillate and their accelerations are augmented with noise to challenge the controller's robustness. 
The driver model parameters are tabulated in Table \ref{table:driver_params}. Note that the parameters of each driver are sampled from uniform distributions, denoted by $\mathcal{U}(\cdot,\cdot)$, to account for heterogeneous behaviors of drivers. Additionally, we assume that all vehicles have the same physical dimensions, for brevity. We also assume that vehicles have perfect perceptions; that is, there is no measurement bias in the vehicles' positions. 

The simulation study utilizes \cite{sisl2019automotive} for simulation and visualization, and runs on a Linux machine (Intel(R) Xeon(R) CPU E5-2640 v4 @ 2.40GHz with NVIDIA GeForce GTX TITAN Black).

\begin{table}[]
\begin{tabular}{@{}lll@{}}
\toprule
\multicolumn{1}{c}{Param} & \multicolumn{1}{c}{Description} & \multicolumn{1}{c}{Value} \\ \midrule
$T$ & Receding time horizon $[s]$ & 2.8\\
$N_{\text{sim}}$ & Control sequence sample size & 32\\
$\Delta t$ & Time step size $[s]$ & 0.4\\
$\lambda_{div}$ & Weight on divergence from a target lane & 12000 \\
$\lambda_{v}$ & Weight on divergence from a desired speed & 1000 \\
$\lambda_{\delta}$ & Weight on steering angle & 500 \\
$\lambda_{a}$ & Weight on acceleration & 500 \\
$\lambda_{\Delta \delta}$ & Weight on steering rate & 100 \\
$\lambda_{\Delta a}$ & Weight on jerk & 100 \\
$\delta_{\text{min}}$ & Minimum steering angle $[rad]$ & -0.3 \\
$\delta_{\text{max}}$ & Maximum steering angle $[rad]$ & 0.3 \\
$a_{\text{min}}$ & Minimum acceleration $[m^2/s]$ & -4.0 \\
$a_{\text{max}}$ & maximum acceleration $[m^2/s]$ & 3.5 \\
$x_{\text{end}}$ & Length of the current lane $[m]$ & 50 \\
$v^{\text{ref}}$ & Desired velocity $[m/s]$ & 10 \\ \bottomrule
\end{tabular}
\caption{\small{Controller design parameters}}
\label{table:control_params}
\end{table}

\begin{table}[]
\begin{tabular}{@{}lll@{}}
\toprule
\multicolumn{1}{c}{Param} & \multicolumn{1}{c}{Description} & \multicolumn{1}{c}{Value} \\ \midrule
$\tilde{v}^{\text{ref}}$ & Reference velocity $[m/s]$ & $\mathcal{U}(2, 5)$ \\
$\tilde{T}$ & Safe time headway $[s]$ & $\mathcal{U}(1, 2)$ \\
$\tilde{a}_{\text{max}}$ & Maximum acceleration $[m/s^2]$ & $\mathcal{U}(2.5, 3.5)$ \\
$\tilde{b}$ & Comfortable deceleration $[m/s^2]$ & $\mathcal{U}(1.5, 2.5)$ \\
$\tilde{\delta}$ & Acceleration exponent & $\mathcal{U}(3.5, 4.5)$ \\
$\tilde{s}_0$ & Minimum distance to front vehicle $[m]$ & $\mathcal{U}(1, 3)$ \\
$\eta_{c}$ & Cooperativeness $\in [0,1]$ & $\mathcal{U}(0, 1)$ \\
$\eta_{p}$ & Perception range $[m]$ & $\mathcal{U}(-0.15, 0.15)$ \\
$w$ & Length from center to side of vehicle $[m]$ & 0.9 \\
$h$ & Length from center to front of vehicle $[m]$ & 2 \\ \bottomrule
\end{tabular}
\caption{\small{Driver model design parameters}}
\label{table:driver_params}
\end{table}

\subsection{Training SGAN for Motion Prediction}
\begin{figure}
    \centering
    \includegraphics[width=1\columnwidth,trim={0cm 0.1cm 0.5cm 1.1cm},clip]{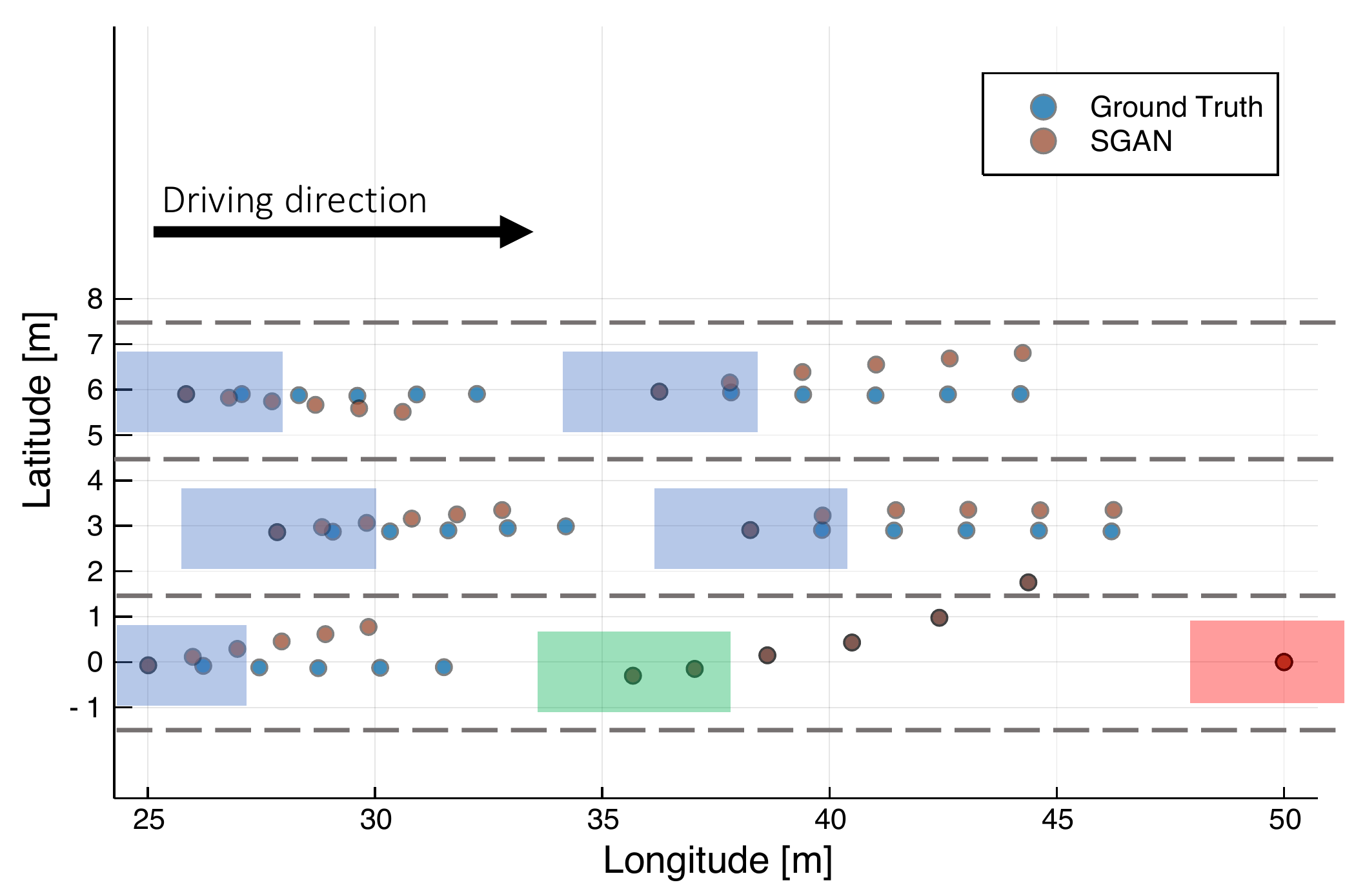}
    \caption{\small{Motions predicted by SGAN. Rectangles in green, blue, and red indicate the ego vehicle, other vehicles, and the stopped vehicles respectively. Dashed lines represent the lane boundary. Each circular point indicates the vehicle center at each time step.}}
    \label{fig:sgan_prediction}
\end{figure}

Table \ref{table:sgan_params} lists hyper parameters for the SGAN used in the simulation study. Training and validation datasets are generated by simulations with only the driver model from Section \ref{ssec:driver_model}, with heterogeneous parameters from Table \ref{table:driver_params}. The dataset is collected from multiple scenarios in various traffic densities, from free flow to dense traffic. In the dataset, we also add noise to the positions so that the trained neural network becomes more robust to noisy inputs. With a total of 27550 data points, training the SGAN with a GPU takes approximately 18 hours. An example of motions predicted by SGAN, compared to ground truth, are illustrated in Fig.~\ref{fig:sgan_prediction}.

The SGAN has 1.872 $[m]$ of average displacement error and 2.643 $[m]$ of final displacement error after $T_{pred}=2$, for the training data. It is an important note that the training dataset does not include the scenario that we test our controller with. This is to examine whether SGAN trained in one environment can still apply to another environment in which SGAN is not trained. 
Note that SGAN can provide a distribution of predicted motions, which can be incorporated into the optimization problem as a chance constraint, thereby enabling a robust formulation. Also, different methods for designing loss functions for SGAN training can be applied, which are topics for future work.

\begin{table}[]
\begin{tabular}{@{}lll@{}}
\toprule
\multicolumn{1}{c}{Param} & \multicolumn{1}{c}{Description} & \multicolumn{1}{c}{Value} \\ \midrule
$T_{obs}$ & Observation time horizon & $8$ \\
$T_{pred}$ & Prediction time horizon & $2$ \\
$N_{b}$ & Batch size & $64$ \\
$D_{emb}$ & Embedding dimension & $64$ \\
$D_{mlp}$ & MLP dimension & $256$ \\
$D_{G,e}$ & Hidden layer dimension of encoder (generator) & $32$ \\
$D_{G,d}$ & Hidden layer dimension of decoder (generator) &  $64$ \\
$D_{D,e}$ & Hidden layer dimension of encoder (discriminator) & $64$ \\
$D_{bot}$ & Bottleneck dimension in Pooling module & $1024$ \\
$\alpha_G$ & Generator learning rate & $5\cdot10^{-4}$ \\
$\alpha_D$ & Discriminator learning rate & $5\cdot10^{-4}$ \\ \bottomrule
\end{tabular}
\caption{\small{SGAN parameters}}
\label{table:sgan_params}
\end{table}

\subsection{Simulation Results and Analysis}
\begin{figure}
    \centering
    \includegraphics[width=1\columnwidth,trim={0.5cm 0.5cm 0.8cm 3.1cm},clip]{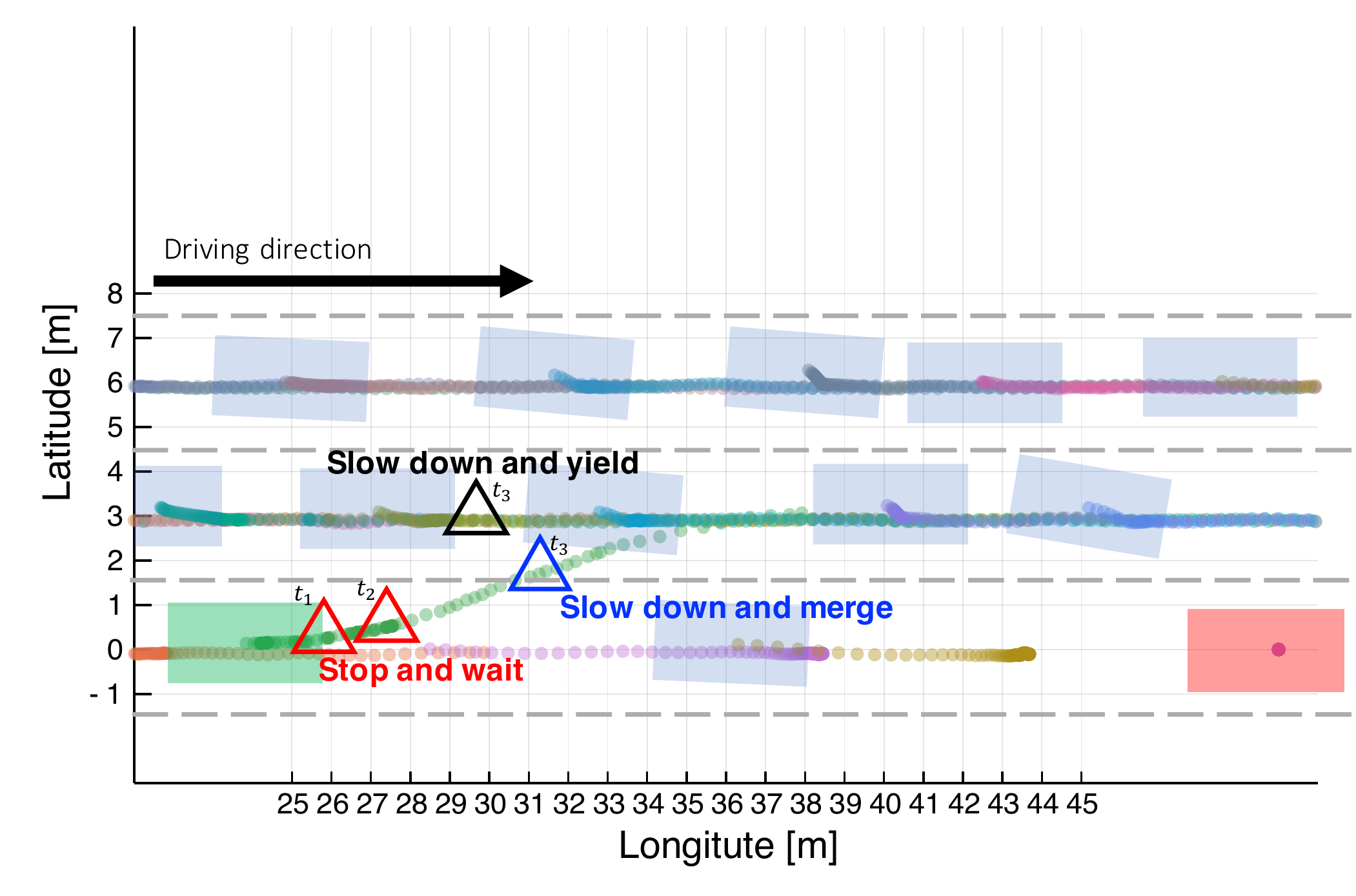}
    \caption{\small{Simulated position trajectories. A dot indicates a center coordinate of a vehicle at each time step and the color of dots distinguishes between trajectories of vehicles. A distance between two successive dots demonstrates how fast a vehicle moves. The triangle labeled with $t_n$ highlights a center position at time $t_n$, $n=1,\ldots,3$. The rectangles illustrate the shape of the vehicles (colors are described in the caption of Fig.~\ref{fig:sgan_prediction}). The positions of the rectangles indicate the initial positions.}}
    \label{fig:sim_positions}
\end{figure}

Fig.~\ref{fig:sim_positions} illustrates the simulated position trajectories. The ego vehicle (in green) often stops and waits before merging (at $t_1$ and $t_2$), since otherwise the safety constraint \eqref{eq:const_collision} can be violated. As the ego vehicle gets closer to the target lane (the middle lane), the vehicle on the target lane reacts by slowing down the speed, to make a space for the ego vehicle to cut in (at $t_3$). As soon as an enough space is made, the ego vehicle merges into the target lane. 

\begin{figure}
    \centering
    \includegraphics[width=1\columnwidth,trim={1.0cm 0.5cm 0.1cm 0.1cm},clip]{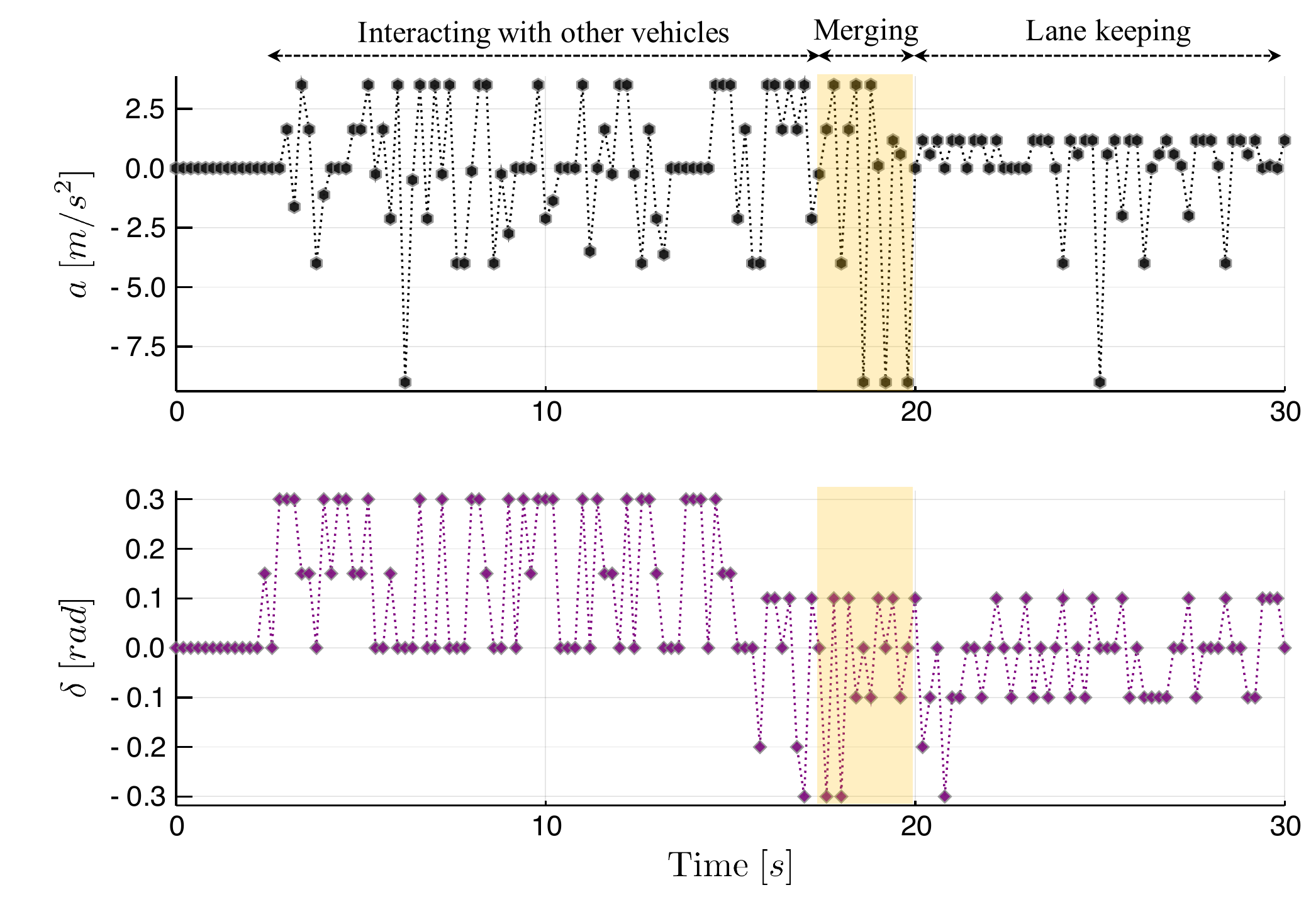}
    \caption{\small{Acceleration (top) and steering angle (bottom) profile of the ego vehicle. The ego vehicle initiates merging to a target lane at time 18.6$[s]$.}}
    \label{fig:sim_controls}
\end{figure}

These interactive behaviors with the ego vehicle are also observed in Fig.~\ref{fig:sim_controls}. While the ego vehicle is interacting with other vehicles to merge over, between 3.2$\sim$18.6$[s]$, both the acceleration and steering angle fluctuate significantly to quickly get closer to the lane while avoiding collision. Between 18.6$\sim$20$[s]$, the ego vehicle changes lanes and the acceleration still fluctuates. This is again because the ego vehicle needs to move quickly, but must stop whenever another vehicle gets too close. Once the ego vehicle merges into the target lane, after 20$[s]$, the acceleration and steering angle fluctuate less, and the vehicle drives smoothly. 

\begin{table}[]
\centering
\begin{tabular}{@{}lllll@{}}
\toprule
 &  & Constant & SGAN & Ground \\ 
 &  & Velocity &  & Truth \\\midrule
Success& Coop. & 96$\%$ & 99$\%$ & 99$\%$ \\
 rate& Mix. & 90$\%$ & 97$\%$ & 97$\%$ \\
 & Agg. & 48$\%$ & 81$\%$ & 89$\%$ \\\midrule
Mean time & Coop. & 25.95 ($\pm 6.42$) & 23.97 ($\pm 5.10$) & 24.53 ($\pm 5.11$) \\
  to merge & Mix. & 26.97 ($\pm 7.07$) & 25.76 ($\pm 5.37$) & 26.21 ($\pm 5.58$) \\
 & Agg. & 35.24 ($\pm 6.12$) & 29.32 ($\pm 7.07$) & 28.31 ($\pm 6.22$) \\\midrule
Mean min  & Coop. & 0.69 ($\pm 0.15$) & 0.49 ($\pm 0.18$) & 0.72 ($\pm 0.14$) \\
 distance& Mix. & 0.58 ($\pm 0.19$) & 0.42 ($\pm 0.19$) & 0.61 ($\pm 0.17$) \\
 & Agg. & 0.35 ($\pm 0.18$) & 0.30 ($\pm 0.14$) & 0.55 ($\pm 0.16$) \\ \bottomrule
\end{tabular}
\caption{\small{Monte Carlo simulation results (a total of 100 simulations each) for three motion prediction models, constant velocity, SGAN, and ground truth, for different level of cooperativeness. A trip is ``successful'' if the ego vehicle changes lanes to the target lane within the time limit (40 [s]). Until the time limit, the target lane is packed with vehicles and there is no empty space where the ego vehicle can overtake without cooperating with other drivers. The ``time to merge'' indicates the time period [s] that the ego vehicle takes to change lanes to the target lane. The ``min distance'' indicates the minimum distance [m] between the ego vehicle and other vehicles at any point of the simulation. The value in the parentheses shows the standard deviation.}}
\label{table:monte}
\end{table}

Note that the other vehicles may or may not be cooperative, depending on their characteristics. Therefore, we quantitatively validate the controller based on Monte Carlo simulations with various characteristics of other drivers. 
In each simulation, all vehicles are randomly positioned, except for the stopped vehicle (dead-end of the road), and the driver model parameters in Table~\ref{table:driver_params} are sampled from uniform distributions. With respect to different cooperation levels, we consider three cases: (i) all drivers are cooperative (i.e. $\eta_{c,i}=1$ for all drivers $i$ in the scene); (ii) all drivers are aggressive (i.e. $\eta_{c,i}=0$); and (iii) they are mixed (i.e. $\eta_{c,i}\in[0,1]$). It is important to highlight that the ego vehicle does not know how cooperative the other drivers are. We found that when the drivers are all cooperative, changing lanes into the target lane within the time limit is more likely than when the drivers are either partially cooperative (i.e. mixed) or non-cooperative (i.e. aggressive), as shown in Table~\ref{table:monte}. Still, it is possible that the ego vehicle cannot change lanes within the time limit (i.e. the success rate is less than 100\%) and ends up being stranded in the current lane. This is because the drivers may have short perception ranges, determined by $\eta_{p}$, which can result in the drivers not detecting the ego vehicle and keeping their speed, even though they are cooperative. Nonetheless, even when all drivers are aggressive, the controller can successfully change lanes in most cases. One caveat is that the ego vehicle can get quite close to the other vehicles, especially when the other drivers are not being cooperative and pass the ego vehicle (mean min distance in Table~\ref{table:monte}), although no collision was observed in any of the simulations. 
%
%
%

We also compare the performance of the controller with SGAN to a simple motion prediction method: a constant velocity model. The constant velocity model predicts that a vehicle will maintain the same speed in the next time step. We also compare the SGAN-enabled controller to a controller with perfect predictions of other vehicles' motions, i.e. ground truth\footnote{In the case of perfect predictions for other vehicles' motion, the simulator propagates other vehicles' motion based on their driver model, and the propagated motions are used by the controller as predictions.}. In general, we found that more accurate predictions lead to higher success rates. 
In fact, when all drivers are cooperative, all three prediction models can lead to successful lane change. That is, the imprecision of predictions on drivers' interactive motions is not critical when the drivers are very cooperative, since the drivers easily submit space to other vehicles, even with rough control inputs resulting from inaccurate motion predictions. This, however, is no longer valid if the drivers are aggressive. When they are aggressive, the ego vehicle needs a precise control that carefully induces cooperation from the other drivers. This cannot occur with the simple constant velocity model. Consequently, the success rate with SGAN is significantly higher (+33$\%$) than that with the simple constant velocity model, when the drivers are aggressive. We also found that the controller with more accurate predictions tends to change lanes more quickly. That is, significant motion prediction errors for the other vehicles lead the ego vehicle to positions where it cannot effectively influence its neighboring vehicle, and create room for a safe lane change. As a consequence, the ego vehicle must let the coming vehicle pass and waits for next vehicle to cooperate. Finally, the minimum distances tend to be significantly larger with perfect predictions compared to that with SGAN. That is, more precise predictions help the controller secure a safer trajectory during lane change. 

Finally, in our recent paper \cite{saxena2019driving}, we compared the performance of the proposed controller with a learning-based controller, in terms of the success rate, time to merge, and minimum distances. The SGAN-enabled controller out-performs the learning-based controller in the success rate, (arguably) safety as measured by minimum distances, and reliability as measured by variances of performance metrics, while taking more time to merge. Interested readers are referred to \cite{saxena2019driving} for detailed comparisons.

Finding a minimum distance between two vehicles \textit{analytically} is computationally efficient, even though it is computed as the minimum distance among any pairs of circles. The circle model takes about $9\cdot10^{-6}[s]$ on average to compute a minimum distance between two vehicles at any time. As a consequence, finding one control input at each time step takes less than $0.2[s]$, which suggests this approach is amenable to \textit{real-time} control. 

Simulation code and videos are available at 
{\tt https://\\github.com/honda-research-institute/\\NNMPC.jl}.

\subsection{Limitations and Future Work}
\label{ssec:limitation_futurework}
No collision was observed in the simulation studies. However, 
the controller cannot guarantee zero collisions with other drivers, especially because in highly dense traffic there is no ``safe'' area that the ego vehicle can merge into without cooperating with other drivers. Still, improved predictions can help the ego vehicle keep a safe distance with other vehicles, as shown in Table~\ref{table:monte}. Improving SGAN prediction accuracy by carefully designing the training set, loss functions, or network structures remains for future work. 

The proposed algorithm somewhat naively generates the control candidates by random sampling, even though action spaces are specified in each driving scenario (keeping lane or changing lane). We can further reduce the actions spaces, using correlation between steering angle and acceleration, for each driving scenario. 

Also, the proposed algorithm (Algorithm \ref{alg:monte_rollout}) is heuristic. That is, even in an identical scenario, the controller may find a different solution. This might be acceptable in many cases where the constraints are not violated. However, an advanced optimization algorithm can be further studied to find a locally optimal solution with a convergence guarantee. 


\section{Conclusion}\label{sec:conclusion}
This paper formalizes a control framework for autonomous lane changing in dense traffic. This paper particularly focuses on heavy traffic where vehicles cannot merge into a lane without cooperating with other drivers. The control framework incorporates a Recurrent Neural Network (RNN) architecture, namely a state-of-the-art Social Generative Adversarial Network (SGAN), to predict interactive motions of multiple drivers. The predicted motions are systematically evaluated in safety constraints to evaluate control inputs. A heuristic algorithm based on Monte Carlo simulation along with a roll-out approach is developed to find feasible solutions in a computationally efficient manner. The qualitative and quantitative analysis in the simulation studies illustrate the strong potential of the proposed control framework for achieving automatic and safe lane changes by cooperating with other drivers.




\bibliography{ref}


\vfill

\end{document}